# A Kronecker product accelerated efficient sparse Gaussian Process (E-SGP) for flow emulation

Yu Duan*, Matthew Eaton, Michael Bluck

Nuclear Engineering Group, Department of Mechanical Engineering, City and Guilds Building (CAGB), Imperial College London, Exhibition Road, South Kensington Campus, SW7 2BX, United Kingdom (UK)

# Abstract

In this paper, we introduce an efficient sparse Gaussian process (E-SGP) for the surrogate modelling of fluid mechanics. This novel Bayesian machine learning algorithm allows efficient model training using databases of different structures. It is a further development of the approximated sparse GP algorithm, combining the concept of efficient GP (E-GP) and variational energy free sparse Gaussian process (VEF-SGP). The developed E-SGP approach exploits the arbitrariness of inducing points and the monotonically increasing nature of the objective function with respect to the number of inducing points in VEF-SGP. By specifying the inducing points on the orthogonal grid/input subspace and using the Kronecker product, E-SGP significantly improves computational efficiency without imposing any constraints on the covariance matrix or increasing the number of parameters that need to be optimised during training.

The E-SGP algorithm developed in this paper outperforms E-GP not only in scalability but also in model quality in terms of mean standardized logarithmic loss (MSLL). The computational complexity of E-GP suffers from the cubic growth regarding the growing structured training database. However, E-SGP maintains computational efficiency whilst the resolution of the model, (i.e., the number of inducing points) remains fixed. The examples show that E-SGP produces more accurate predictions in comparison with E-GP when the model resolutions are similar in both. E-GP benefits from more training data but comes with higher computational demands, while E-SGP achieves a comparable level of accuracy but is more computationally efficient, making E-SGP a potentially preferable choice for fluid mechanic problems. Furthermore, E-SGP can produce more reasonable estimates of model uncertainty, whilst E-GP is more likely to produce over-confident predictions.

* Corresponding author: y.duan@imperial.ac.uk

# 1 Introduction

Fluid mechanics is pivotal to industries as diverse as transport, power generation, pharmaceutical industries, and more. Physics-based approaches are essential for understanding complex, three-dimensional (3D) turbulent thermo-fluid phenomena. However, it remains challenging to use a physics-based methodology to efficiently address the challenges posed in performing extensive parametric CFD simulations, e.g., uncertainty quantification (UQ), design exploration and optimisation. There is no doubt that the analysis of complex 3D turbulent thermo-fluid behaviour within complex engineering systems is a computationally demanding field of engineering involving very large data sets. This makes it an ideal field of engineering application to apply machine learning (ML) algorithms for producing surrogate models as well as performing design optimisation, data mining and anomaly or fault detection [1,2]. The successful, and disruptive, impact of ML algorithms in applications such as language and image processing [3], advertising [4], and finance [5], has attracted lots of attention from the fluid mechanics community. Developing interpretable, generalisable, and robust ML algorithms has become one of the goals of ML research related to fluid mechanics. Many reviews have been published recently on the development and application of ML in fluid mechanics, e.g. [6–8]. In this paper, we propose an efficient analytic Bayesian ML algorithm for learning the complex flow dynamics from structured and unstructured computational mesh databases.

## 1.1 Why Bayesian?

The success of deterministic ML models, e.g., deep neural networks (DNNs), is often plagued by several issues, such as overfitting due to small or noisy datasets [9]. Both computational and experimental fluid mechanics often produce very large data sets, or quantities of interest (QoI), such as temperature, pressure, velocity and other associated thermo-fluid data. However, the production of high-fidelity computational and experimental data can be costly to produce in terms of computational resources as well as financially in terms of the cost of thermo-fluid experiments. Consequently, the amount of high-fidelity thermo-fluid data available may be small compared to the complexity of the problem being analysed. Furthermore, the lack of reliable confidence estimates and other robustness issues in deterministic ML makes it vulnerable to adversarial attacks [10] and also reducing the models' reproducibility [11]. As we know, the confidence/uncertainty quantification (UQ) of model predictions is essential for reliability, and risk assessment, in safety-critical industries [12]. Finally, prior knowledge of fluid mechanic problems (e.g., the governing equations or empirical knowledge) are often available. Therefore, this makes it an attractive field of research for the application of Bayesian ML algorithms.

## 1.2  Motivation for the Development and Use of Gaussian Process based Bayesian ML Algorithms

Gaussian process (GP) based methods represent a set of powerful analytical Bayesian ML algorithms. It assumes that any combination of observations of a complex system forms an adjoint multivariate Gaussian distribution. The key to GP model training is to find the appropriate covariance matrix that maximises the marginal likelihood. The predictions of the GP model can then be interpreted as using existing knowledge to calculate the conditional distribution of function outputs at unknown points. This conditional distribution also has the form of a Gaussian distribution, as such the prediction of the GP model consists of two parameters, the mean, and the variance. The variance can be used to calculate confidence intervals of model predictions. Even more attractive, is the fact that the entries in the covariance matrix are computed using the covariance kernel function (kernel function for short), which can be re-derived from the PDF of the problem that is being analysed. This enables the integration of the governing physics laws within the algorithm, e.g. Latent force models (LFM) [13–17], numerical GP [18] and GP for linear [19] and non-linear PDFs [20]. (For the sake of clarity, we will name the original definition of GP as standard GP or STD-GP in the following discussion.)

## 1.3  Background

Covariance matrix inversion and storage are two major obstacles to the general application of Gaussian processes (GP) to engineering and thermo-fluid problems involving very large data sets. For a standard GP model with N training data points, the computational complexity is $\mathcal{O}(N^3)$ and the storage requirement is $\mathcal{O}(N^2)$. For a problem with more than $10^4$ training data points, STD-GP becomes impractical. Many efforts have been made to solve this problem. Various scalable GPs have been developed to improve the scalability of the method. Recently, Liu et al. [21] conducted a comprehensive review of the scalable GP, in which they classified the scalable GPs into two major categories: 1) global approximation and 2) local approximation. The global approximation is further divided into 1.1) subset of data, 1.2) sparse kernel, 1.3) sparse approximation. The work in this paper is consistent with the sparse approximation scalable GPs.

To improve the scalability of the STD-GP, the sparse approximate inference GP method adopts the concept of the inducing inputs e.g. [22–25]. The inducing points is a set of representative points being used to summarise the information from the data. Titsias [26] proposed variational inducing points sparse GP or VEF-SGP (the name of this approach refers to [21]). VEF-SGP treats the outputs of a function at inducing points as latent variables and derives a procedure for approximating latent variables and hyperparameters. The method reduces the risk of over-fitting by introducing an extra regulation term in the objective function. Hensman et al. [27,28] scales up the VEF-SGP using stochastic variational inference (SVI) and generalized the approach to classification problems. Although sparse approximate inference GP methods greatly improve the efficiency of STD-GP, the limitation is also obvious. The number of inducing point is usually restricted to a number smaller than $\mathcal{O}(10^4)$, which again may not be enough for a complex and high-dimensional problem.

The STD-GP can also be accelerated by exploiting the structure of the training dataset. Saatçi [29] proposed an efficient GP (E-GP) that utilises kernel functions in the form of tensor products and a structured training dataset. In E-GP, the covariance matrix is calculated as the Kronecker product of the sub-covariance matrices for each input. Similar approaches have been applied to accelerate the sparse approximate inference GP approaches. Wilson and Nickisch [30] proposed KISS-GP which uses the structured kernel interpolation (SKI) to approximate the covariance matrix. A development of KISS-GP for the product kernel function can be found in [31], while the online version of the KISS-GP, WISKI-GP was developed by Stanton et al. [32]. The scalability of the SKI-GP method is further improved in [33]. Undoubtedly, the adoption of the SKI approaches greatly improves the time efficiency of GP model training, but it may also lead to discontinuity and overconfidence in model prediction [21], which are unfavourable for fluid mechanics problems.

Izmailov, Noikov and Kropotov [34] proposed the TT-GP which uses the tensor train decomposition for the high-dimensional dataset, and also assumes that the covariance matrix of the inducing points can always be factorised using a Kronecker product. However, this assumption restricts the covariance matrix representation [35] and also increases the number of parameters that need to be optimised during training (e.g. entries in the sub-covariance matrices of inducing points). Atkinson and Zabaras [36] proposed an unsupervised learning algorithm based on the Bayesian GP-LVM for structured datasets. This structured, GP-LVM (SGP-LVM) algorithm, is computationally tractable and achieves a complexity similar to the SKI-GP algorithm without placing strict constraints on the covariance matrix or greatly increasing the number of parameters.

In this paper, we will derive a computationally tractable and efficient VEF-GP method in a similar manner to that described in [36]. In the following context, we refer to this novel VEF-GP approach as E-SGP. The rest of this paper is organized as follows. In Section 2, the mathematics of STD-GP, VEF-GP, as well as the E-GP are reintroduced, while mathematical description of the E-SGP is included in Section 3. Applications of the E-SGP with spatially structured and unstructured mesh fluid dynamics datasets are discussed in Section 4, where we also compare the quality and scalability of E-SGP to those of E-GP. Section 5 summaries our work and discusses the potential of E-SGP

## 2  STD-GP, VEF-GP and E-GP revisited

In this section, we begin with a mathematical description of the STD-GP followed by a concise description of the variational learning sparse GP (VEF-GP) [26] and efficient GP formula [29]. This allows us to introduce notation and derive expressions so that the Kronecker product accelerated efficient SGP (E-SGP) can be formulated. If readers require more information on GPs, the following references are recommended [21,24,37–39].

### 2.1   Gaussian processes (STD-GP) revisited.

In standard GP (STD-GP), we assume that any finite subset of $\{f(\vec{x})|\vec{x} \in \mathbb{R}^D\}$ follows a multivariate Gaussian distribution. Considering the vector of noisy observations $\vec{y} = \{y_i \in \mathbb{R}\}_{i=1}^n$ of the latent function acquired at $n$ locations $X_n = \{\vec{x}_i \in \mathbb{R}^D\}_{i=1}^n$, $\vec{y}$ have the priors given by

$$p(\vec{y}|\vec{f}, X_n) = \mathcal{N}(\vec{y}|\vec{f}, \sigma^2 I_{nn}), \qquad (1)$$

$$p(\vec{f}|X_n) = \mathcal{N}(\vec{y}|\vec{m}(X_n), K_{nn}). \tag{2}$$

In Eq. (1), $\vec{f}$ is the function outputs at locations $X_n$. $\vec{m}(X_n)$ denotes the mean function values at $X_n$. For the sake of simplicity, $\vec{m}(X_n)$ can be fixed to zeros in many GP applications. $K_{nn} = K(X_n, X_n)$ in Eq. (2) is an $n \times n$ real symmetric positive semi-definite matrix, whose elements are evaluated using kernel functions ($k(x, x')$). $\sigma$ represents the noise/error level of $\vec{y}$ and $I_{nn}$ is an n-by-n identical matrix.

The training of the GP data-driven model means optimising the hyperparameters of the kernel function to maximise the logarithmic marginal likelihood $p(\vec{y}|X_n)$, which is written as

$$\log p(\vec{y}|X_n) = -\frac{n}{2}\log 2\pi - \frac{1}{2}\vec{y}(\sigma^2 I_{nn} + K_{nn})^{-1}\vec{y}^T - \frac{1}{2}\log|\sigma^2 I_{nn} + K_{nn}|, \tag{3}$$

or using the eigen-decomposition of the covariance matrix

$$\log p(\vec{y}|X_n) = -\frac{n}{2}\log 2\pi - \frac{1}{2}\vec{y}V_{nn}(\sigma^2 I_{nn} + E_{nn})^{-1}V_{nn}^T\vec{y}^T$$
$$-\frac{1}{2}\log|\sigma^2 I_{nn} + E_{nn}|. \tag{4}$$

In Eq. (4), $V_{nn}$ is the square $n \times n$ matrix whose columns are the eigenvectors of $K_{nn}$, and $E_{nn}$ is the diagonal matrix and its diagonal elements are the corresponding eigenvalues.

Based on the prior assumption of the STD-GP, we can write the joint distribution of the unknown function values $\vec{f}_*$ at the input locations $X_* = \{\vec{x}_i^* \in \mathbb{R}^D\}$ and noisy observation $\vec{y}$ as

$$\begin{bmatrix} \vec{y} \\ \vec{f}_* \end{bmatrix} \sim N\left(\vec{0}, \begin{matrix} K_{nn} + \sigma^2 I_{nn} & K(X_*, X_n)^T \\ K(X_*, X_n) & K(X_*, X_*) \end{matrix}\right), \tag{5}$$

which also lead to the predictive equations of the STD-GP model:

$$\langle \vec{f}_* \rangle = K(X_*, X_n)(\sigma^2 I_{nn} + K_{nn})^{-1}\vec{y}^T, \tag{6}$$

$$cov(\vec{f}_*) = K(X_*, X_*) - K(X_*, X_n)(\sigma^2 I_{nn} + K_{nn})^{-1}K(X_*, X_n)^T. \tag{7}$$

where $\langle \vec{f}_* \rangle$ is the expected values of predictions, and $cov(\vec{f}_*)$ is the covariance matrix of the predictions providing information about how predictions at different points are correlated, with the diagonal elements representing the variances of the predictions at those points.

## 2.2 Efficient Gaussian Process (E-GP)

Saatçi [29] proposed an efficient GP (E-GP) algorithm for structured datasets. More specifically, the data are on a multidimensional Cartesian grid $X = X_1 \times X_2 \times \cdots \times X_n$, and the kernel function has a tensor product form. In particular, $k(\cdot,\cdot)$ is a tensor product if it can be written as $k(\vec{x}, \vec{x}') = k(x_1, x_1')k(x_2, x_2') \cdots k(x_n, x_n')$. For inputs defined on X, the covariance matrix $K_{nn}$ can then be written in the form of a Kronecker product:

$$K_{nn} = \otimes_{i=1}^{n} K_i, \qquad (8)$$

where $K_i$ is the square covariance matrix of the inputs on the $X_i$ axis. For the sake of simplicity, $K_i$ is named as sub-covariance matrix below. The eigen-decomposition of $K$ can now be expressed as

$$K_{nn} = VDV^T = \otimes_{i=1}^{n}(V_i E_i V_i^T) = (\otimes_{i=1}^{n} V_i)(\otimes_{i=1}^{n} E_i)(\otimes_{i=1}^{n} V_i^T), \qquad (9)$$

and $K_{nn}^{-1} = (\otimes_{i=1}^{n} V_i)(\otimes_{i=1}^{n} E_i^{-1})(\otimes_{i=1}^{n} V_i^T)$

In Eq. (9), $E_i$ is a diagonal matrix containing the eigenvalues of the $K_i$.

Eq. (3) is now written as

$$\log p(\vec{y}|X_n) = -\frac{n}{2}\log 2\pi - \frac{1}{2}\vec{y}(\otimes_{i=1}^{n} V_i)(\sigma^2 I + \otimes_{i=1}^{n} E_i)^{-1}(\otimes_{i=1}^{n} V_i^T)\vec{y}^T \qquad (10)$$
$$-\frac{1}{2}\log|\sigma^2 I + \otimes_{i=1}^{n} D_i|.$$

The computational complexity of $(\otimes_{i=1}^{n} V_i^T)\vec{y}^T$ can be reduced to $\mathcal{O}(N)$ using the mixed Kronecker matrix-vector product:

$$(B^T \otimes A)\vec{y}^T = vec(AYB), \qquad (11)$$

in which $vec$ is the vectorization operator e.g. $Y = vec(\vec{y}^T)$ and $\vec{y}^T$ is a column vector. The computational complexity for evaluating marginal likelihood represented by Eq. (4) is now reduced to $\mathcal{O}(N \sum_{i=1}^{n} n_i)$, where $N = \prod_{i=1}^{n} n_i$. The similar approach is also applicable to speed up the prediction.

E-GP can be easily extended to problems where the input space consists of distinct and independent subspaces. In other words, $X_i$ in X is not only interpreted as inputs on an axis but also inputs in an independent subspace. This will ease the restriction of E-GP on the structure of the dataset and is especially useful in fluid mechanics problems. For instance, the inputs of a fluid mechanics problem can be easily divided into several orthogonal subspaces, like spatial inputs, time inputs, operational parameters, design parameters, etc. But the obvious limitation of E-GP (the number of inputs on each grid/input subspace must be smaller than $\mathcal{O}(10^4)$) will again hinder its generality in recovering the complex flow field. In real applications, the number of spatial inputs is normally more than $\mathcal{O}(10^4)$.

## 2.3 Variational energy free sparse Gaussian process (VEF-SGP)

Let us define a set of inducing variables $\vec{u} = \{u_i\}_{i=1}^m$ defined at $X_m = \{\vec{x}_i \in \mathbb{R}^D\}_{i=1}^m$. And we also have $m \ll n$. Follow the assumption of GP, the distribution of $\vec{u}$ with knowledge of $\vec{y}$ can be written as $p(\vec{u}|\vec{y})$). To reduce the overfitting risk, the VEF-SGP algorithm introduces variational distribution of $\vec{u}$, $q(\vec{u}) \sim N(\vec{u}|\vec{\mu}, S_{mm})$ (In other words, $q(\vec{u})$ is the variational distribution of $p(\vec{u}|\vec{y})$). $X_m, \vec{\mu}$ and $S_{mm}$ together is called as variational quantities. Now, the posterior GP mean and variance for latent function at the unknow location $\vec{x}^*$ w.r.t $q(\vec{u})$ as

$$\bar{y}^* = K_{x^*m} K_{mm}^{-1} \vec{\mu}, \tag{12}$$

$$var(y^*) = k(\vec{x}^*, \vec{x}^*) - K_{x^*m} K_{mm}^{-1} K_{mx^*} + K_{x^*m} K_{mm}^{-1} S_{mm} K_{mm}^{-1} K_{mx^*}, \tag{13}$$

where $var(\cdot)$ is the variance of model prediction.

Except for $\vec{\mu}, S_{mm}$ and the hyperparameters, the quality of the approximation SGP is also dependent on the locations of the subset-of-data or inducing points in VEF-SGP. Many methods proposed to optimise the location of the inducing point with respect to posterior likelihood [25,40,41], information gain[25] or prediction error[42]. But it will not present as a big issue, since EBLO ($\ell$) will converge to the value of the STD-GP model with the full training dataset as the number of inducing points increases[26]. As such it is reasonable to say that inducing points in VEF-SGP can be treated as the mesh in computational fluid dynamics (CFD) simulations. And, the effect of inducing points can be minimised using the sensitive study akin to that in the CFD simulations. In this paper, we always treat the location of the inducing points as user defined.

In VEF-SGP, $\vec{\mu}, S_{mm}$ and the hyperparameters are determined by maximising the evidence lower boundary (ELBO) of the logarithm marginal likelihood. The EBLO is written as

$$\ell = -\frac{n}{2}\log(2\pi) - \frac{1}{2}\vec{y}(\sigma^2 I_{nn} + K_{nm} K_{mm}^{-1} K_{mn})^{-1} \vec{y}^T \tag{14}$$

$$-\frac{1}{2}\log|\sigma^2 I_{nn} + K_{nm} K_{mm}^{-1} K_{mn}| - \frac{1}{2\sigma^2} tr(\widetilde{K}),$$

where $\widetilde{K} = K_{nn} - K_{nm} K_{mm}^{-1} K_{mn}$ and $tr(\widetilde{K}) = tr(K_{nn}) - tr(K_{mm}^{-1} K_{mn} K_{nm})$. The optimal parameters of $q(\vec{u})$ ($\vec{\mu}$ and $S_{mm}$) are written as:

$$S_{mm} = K_{mm}(K_{mm} + \sigma^{-2}K_{mn}K_{nm})^{-1}K_{mm} \text{ and } \vec{\mu} = \sigma^{-2}S_{mm}K_{mm}^{-1}K_{mn}\vec{y} \quad (15)$$

It is worth noting that the 4$^{\text{th}}$ term in Eq. (10), $\frac{1}{2\sigma^2} tr(\widetilde{K})$, is the regularisation term which reduce the overfitting risk and ensure the monotonically increasing feature of $\ell$ as the number of inducing increases.

Both $K_{mm}$ and $K_{mm}^{-1}$ are the real symmetric positive definite matrices, therefore $K_{mm}^{-1}$ can be represented as the production of the low triangular matrix ($L_{mm}$) and its transpose using Cholesky decomposition. Hereafter we write $K_{nm}K_{mm}^{-1}K_{mn} = Q_{nm}Q_{mn}$, where $Q_{nm} = K_{nm}L_{mm}$ and $Q_{mn} = Q_{nm}^T$. Note that $Q_{mn}Q_{nm}$ is also real and symmetric and thus guarantees the orthogonality of its eigen vectors. The eigen decomposition of $Q_{mn}Q_{nm}$ can then be written as $W_{mm}E_{mm}W_{mm}^T$, while $E_{mm}$ is the diagonal matrix containing eigenvalues and the columns of $W_{mm}$ are the eigenvectors of $Q_{mn}Q_{nm}$. Based on the matrix inversion lemma, also known as Woodbury formula, we now have:

$$(\sigma^2 I_{nn} + K_{nm}K_{mm}^{-1}K_{mn})^{-1} = (\sigma^2 I_{nn} + Q_{nm}Q_{mn})^{-1} \quad (16)$$
$$= \sigma^{-2}I_{nn} - \sigma^{-2}Q_{nm}W_{mm}(\sigma^2 I_{mm} + E_{mm})^{-1}W_{mm}^T Q_{mn},$$
and
$$\log|\sigma^2 I_{nn} + K_{nm}K_{mm}^{-1}K_{mn}| = (n-m)\log \sigma^2 + \log|\sigma^2 I_{mm} + E_{mm}|.$$

After submitting Eq. (16) into Eq. (14), we have the new form of the ELBO

$$\ell = -\frac{n}{2}\log(2\pi) - \frac{1}{2}\sigma^{-2}\vec{y}\vec{y}^T - \frac{1}{2}(n-m)\log\sigma^2 \quad (17)$$
$$+ \frac{1}{2}\sigma^{-2}\vec{y}Q_{nm}W_{mm}(\sigma^2 I_{mm} + E_{mm})^{-1}W_{mm}^T Q_{nm}^T \vec{y}^T$$
$$- \frac{1}{2}\log|\sigma^2 I_{mm} + E_{mm}| - \frac{1}{2\sigma^2}tr(\widetilde{K}).$$

The optimal parameters of $q(\vec{u})$ ($\vec{\mu}$ and $S_{mm}$) in [26] can also be re-written as:

$$S_{mm} = K_{mm} - \sigma^{-2}K_{mn}K_{nm} \quad (18)$$
$$+ \sigma^{-2}K_{mn}Q_{nm}W_{mm}(\sigma^2 I_{mm} + E_{mm})^{-1}W_{mm}^T Q_{mn}K_{nm},$$
$$\vec{\mu} = \sigma^{-2}[K_{mn} - \sigma^{-2}K_{mn}Q_{nm}Q_{mn} \quad (19)$$
$$+ \sigma^{-2}K_{mn}Q_{nm}W_{mm}(\sigma^2 I_{mm} + E_{mm})^{-1}W_{mm}^T Q_{mn}Q_{nm}Q_{mn}]\vec{y}.$$

## 3 Kronecker product accelerated VEF-SGP (E-SGP)

In this section, we combine concepts of E-GP and VEF-SGP to develop the Kronecker product accelerated SGP (E-SGP) algorithm. Here we always assume that the studied problem is defined in an input space what can be divided into multiple orthogonal subspaces. Suppose the inputs of the observations $\vec{y}$ is $\chi^1 \times \chi^2 \times \cdots \times \chi^s, \chi^i \in R^{d_i}, i = 1, \cdots$, where, $\times$ denotes the Cartesian product between subsets. $R^{d_i}$ is an input subspace with dimension $d_i$, and the total dimension of the input space $d = \sum_{i=1}^{s} d_i, i = 1, \cdots, s$. $\chi^i$ is a $n_i \times 1$ vector if $d_i = 1$ or a $n_i \times d_i$ matrix otherwise, where $n_i$ is number of inputs in $\chi^i$ and the number of observations $n = \prod_{i=1}^{s} n_i$. We also define the inducing points using the form $\chi_u^1 \times \chi_u^2 \times \cdots \times \chi_u^s$. $\chi_u^i \in R^{d_i}, i = 1, \cdots, s$ and $m = \prod_{i=1}^{s} m_i$, while $m_i$ is the number of inducing points in $\chi_u^i$. Since the inducing point can be arbitrary, for the sake of the computational simplicity, we further assume that inputs in any $\chi_u^i$ always locate on a cartesian grid, namely $\chi_u^i = X_u^{i,1} \times X_u^{i,2} \times \cdots X_u^{i,d_l}$.

Like E-GP, the kernel functions with the tensor product format will also be used here. Thence, $Q_{nm}$ and $Q_{mn}Q_{nm}$ in the Eq. (16) can be expressed as

$$Q_{nm} = K_{nm}L_{mm} = \otimes_{i=1}^{s} \left( K_{nm}^i L_{mm}^i \right), \text{ and} \tag{20}$$

$$Q_{mn}Q_{nm} = \otimes_{i=1}^{s} \left( Q_{mn}^i Q_{nm}^i \right) = \otimes_{i=1}^{s} \left[ W_{m_i m_i} E_{m_i m_i} W_{m_i m_i}^T \right]$$
$$= \left( \otimes_{i=1}^{s} W_{m_i m_i} \right) \left( \otimes_{i=1}^{s} E_{m_i m_i} \right) \left( \otimes_{i=1}^{s} W_{m_i m_i}^T \right).$$

In Eq. (20), $K_{nm}^i$ is the covariance matrix between inputs in $\chi^i$ and inducing inputs in $\chi_u^i$ while $L_{mm}^i$ is the triangular matrix obtained using Cholesky decomposition for the covariance matrix of inducing inputs in $\chi_u^i$. In addition, the regulation term $tr(\widetilde{K})$ in Eq. (17) can now be calculated as

$$tr(\widetilde{K}) = \prod_{i=1}^{s} \left( tr(K_{nn}^i) \right) - \prod_{i=1}^{s} \left( tr\left( (K_{mm}^i)^{-1} K_{mn}^i K_{nm}^i \right) \right). \tag{21}$$

Hereby, the computational complexity of Eq. (17) is now reduced to $\mathcal{O}(N \sum_{i=1}^{n} m_i^2)$ and is much more efficient than E-GP when $m_i \ll n_i$. Eq. (17) can be furtherly accelerated as we enforce the inducing points on a Cartesian grid in each input subspace, since $K_{mm}$ can be rewritten in the form of Kronecker product with respect to each input.

The approximated GP mean and variance in Eq. (12 & 13) can be expend as follows:

$$\bar{y}^* = \sigma^{-2} B_{x^*n} \vec{y} - \sigma^{-4} B_{x^*n} Q_{nm} Q_{mn} \vec{y} \qquad (22)$$
$$+ \sigma^{-4} B_{x^*n} Q_{nm} W_{mm} (\sigma^2 I_{mm} + E_{mm})^{-1} W_{mm}^T Q_{mn} Q_{nm} Q_{mn} \vec{y},$$
$$var(y^*) = k(\vec{x}^*, \vec{x}^*) - \sigma^{-2} B_{x^*n} B_{x^*n}^T \qquad (23)$$
$$+ \sigma^{-2} B_{x^*n} Q_{nm} W_{mm} (\sigma^2 I_{mm} + E_{mm})^{-1} W_{mm}^T Q_{mn} B_{x^*n}^T,$$

where $B_{x^*n} = K_{x^*m} K_{mm}^{-1} K_{mn}$.

Once the unknown inputs has the similar form as the training data, $B_{x^*n}$, $B_{x^*n} Q_{nm} Q_{mn}$, an $B_{x^*n} Q_{nm} W_{mm}$ in Eq. (22) can be written in the Kronecker product form. The first two term in Eq. (22) can be effectively evaluated using Eq. (11). For the third term in Eq. (22), $(\sigma^2 I_{mm} + E_{mm})^{-1} W_{mm}^T Q_{mn} Q_{nm} Q_{mn} \vec{y}$ is eventually a $m \times 1$ vector and can be easily calculated suing Eq. (11) and element wise production between a column vectors $(diag((\sigma^2 I_{mm} + E_{mm})^{-1}))$ and $m \times 1$ vector $W_{mm}^T Q_{mn} Q_{nm} Q_{mn} \vec{y}$. Similar method can also be applied to accelerate the calculation of GP variance.

## 4 Experiments

Both E-GP and E-SGP algorithms are used to learn the periodic bluff body flow and the lid-driven cavity flow generated using CFD with different types of meshing strategies. In the first example, the periodic bluff-body flow field is generated using a structured grid (section 4.1). The structured database allows both algorithms to take full advantage of the Kronecker product. In the second example, E-GP and E-SGP models are trained to learn the relationships between the moving wall and the flow field in a square cavity (Section 4.2). The quality and scalability of both algorithms will be evaluated on the database with unstructured spatial inputs. The last experiment in section 4.3 aims to test the quality and scalability of both algorithms on the database generated using different meshes and again using lid-drive cavity flow as an example. Both E-GP and E-SGP algorithms are implemented in our in-house MATLAB code. All experiments are performed on a workstation with Intel XI(R) E5-2687w 3.0GHz CPU and 254 GB RAM.

### 4.1 Transient streamwise velocity behind a prism bluff body.

In this section, E-GP and E-SGP are applied to learn the transient velocity field behind a prism bluff body. The cross-section of the prism bluff body is an equilateral triangle with side length $a = 0.04m$. The prism bluffy body is in a channel with height, length, and depth of $3a$, $28a$, and $8a$, respectively. The cross-section of the flow channel and prism bluff body is depicted in Figure 1, where the x-axis represents the flow direction, and the y-axis represents the normal direction of the wall. The flow Reynolds number ($Re_a$) calculated from the side length (a) is 100. The presence of the prism bluff body causes a large-scale flow separation, leading to unsteady flow and periodic asymmetrical vortex shedding.

The training data is generated using the unsteady laminar flow solver and structured mesh grid in the commercial CFD software STAR-CCM+. The time step interval in the CFD model is set as 0.01s, while the grid resolution in the x, y, and z directions is $157 \times 60 \times 80$. The z-axis is perpendicular to the xy-plane. For simplicity, we only consider the fluctuating component of the streamwise velocity ($u_x$) recorded between 0 and 10a in the xy-plane, see the red rectangular in Figure 1. Also, the grey dots (·) in Figure 1 (b) represent the spatial locations where the training data is sampled. The temporal resolution of the training data is ten times (0.1s) the time step interval (0.01s) in the CFD model, whilst the time span of the recorded training data is 15s.

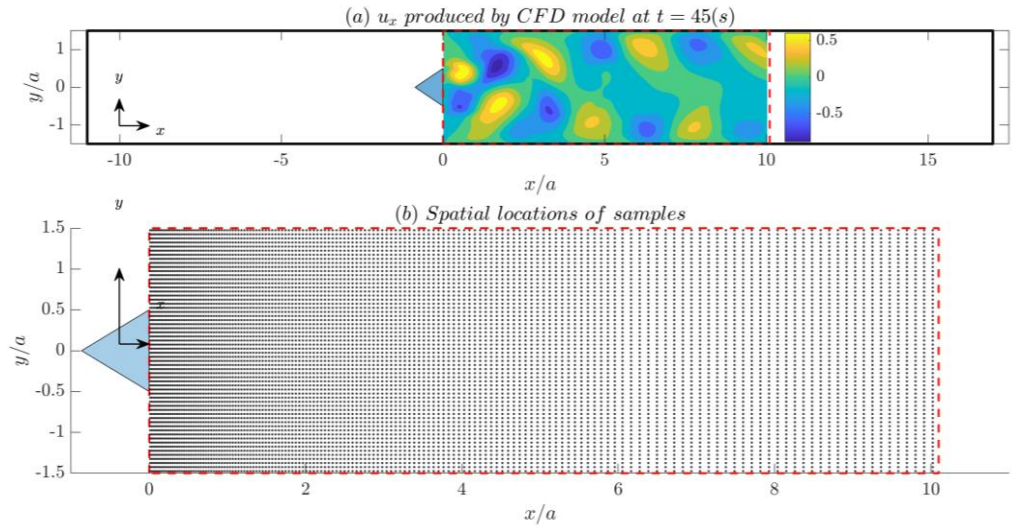

Figure 1 (a) Cross section of the flow domain, the considered sampling range of training data for data-driven models, as well as a single snapshot of the streamwise fluctuating velocity component ($u_x$), (b) spatial locations of samples.

Before constructing the kernel function for GP models of the transient flow domain, we assume the quantity-of-interest has the form of $f(\vec{x}) = \prod_{i=1}^{d} f_i(x_i)$, where $d$ is the dimension of the input space or the number of independent subspace. The kernel function for the latent process $f(\vec{x})$ can then be expressed as $k(\vec{x}, \vec{x}') = \prod_{i=1}^{d} k_i(x_i, x_i')$. Here, the fluctuating component of the streamwise velocity, $u(x, y, t)$, is assumed to be the product of three independent functions $u_1(x), u_2(y), and\ u_3(t)$. And the kernel function $k(x, y, t, x', y', t')$ for $u(x, y, t)$ can then be written as $k_x(x, x')k_y(y, y')k_T(t, t')$. In the example, the squared exponential (SE) kernel function is adopted for the spatial inputs and the periodic kernel is used for the temporal inputs. Hereby, the kernel function for $u(x, y, t)$ can be written as

$$k\big((x, y, t), (x', y', t')\big) \tag{24}$$
$$= \sigma_f^2 \exp\left(-\frac{(x - x')^2}{2l_x^2}\right) \exp\left(-\frac{(y - y')^2}{2l_y^2}\right) \exp\left(-\frac{1}{2w_a^2} \sin^2\left(\pi \left|\frac{t - t'}{\tau}\right|\right)\right).$$

Figure 2 shows snapshots of $u_x$, predicted using the physics-based and data-driven methods, as well as errors in the data-driven model predictions. It is worth mentioning that the number of data points in the E-GP model is the same as the number of inducing points in the two E-SGP models. Specifically, 13 points are assigned in the y-direction and 33 points in the x-direction. Furthermore, the temporal inputs in all three data-driven models is the same as that of the training data.

Figure 2 (a) is the CFD predictions of the $u_x$ field downstream of the prism bluff body at t=45s, which is 15s away from the training dataset. The predictions made by the trained E-GP model with subset of data (SoD-E-GP), E-SGP model with the previous SoD (SoD-E-SGP) as the inducing input and E-SGP with uniformed distributed inducing inputs (Xu-E-SGP) are shown in Figure 2 (b, c, and d). Figure 2 (e, f, and g) shows errors in the prediction of the related data-driven models. As illustrated in Figure 2, the accuracy of the SoD-E-GP model (Figure 2(b & e)) and the SoD-E-SGP model (Figure 2 (c & f)) deteriorates towards the outlet. This is because both SoD-E-GP and SoD-E-SGP inherit the non-uniform nature of the training data in the x-direction. The spatial resolution of both models becomes coarser toward the outlet. By evenly distributing the inducing points, the Xu-E-SGP model reduces the error significantly, see Figure 2 (d and g). Finally, the accuracy of the SoD-E-SGP is slightly improved in comparison with the SoD-E-GP. Later, the statistical performance of E-GP and E-SGP due to varying input locations will be discussed.

To create a baseline, predictions of CFD models with different structured mesh resolutions are used to assess errors due to varying mesh densities. Similar resolutions are then applied to selecting subsets of data (SoD) for the E-GP model and determining the inducing points (Xu) for the E-SGP model. A total of five different spatial resolutions (in the xy-plane) are adopted, namely '8×21', '13×33', '16×41', '21×80' and '31×95'. Spatial coordinates of SoD or Xu are chosen randomly using the 'randperm' or 'lhsdesign' functions in MATLAB. This process is repeated a hundred times, giving rise to the statistics of the accuracy of the data-driven models which are shown in Figure 3. The root mean square error (RMSE) defined in Eq.(25) and mean standardised log loss (MSLL) defined in Eq. (26) of both E-GP and E-SGP gradually reduces as more data points/inducing points (model resolutions in other words) are included in the model. For the same data input resolution, the RMSE and MSLL of the E-SGP model are likely smaller than those of the E-GP model. In particular, E-SGP evidently surpass E-GP when the data points/inducing points are much sparser than the resolution of the training data, i.e. '8 × 21'.

$$RMSE = \left(\frac{1}{n_t}\sum_{i=1}^{n_t}(y_i^* - \bar{y}_i^*)^2\right)^{1/2}, \qquad (25)$$

$$MSLL = \frac{1}{n_t}\left\{\sum_{i=1}^{n_t}\left[\frac{1}{2}\log(2\pi var(y_i^*)) + \frac{(y_i^* - \bar{y}_i^*)^2}{2var(y_i^*)}\right]\right\}. \qquad (26)$$

The minimum RMSE and associated MSLL for E-GP and E-SGP models are plotted in Figure 4. For comparison purposes, the RMSE of the coarse mesh CFD model, the metrics of the E-GP model with full dataset, as well as the E-SGP model with the same spatial inputs of the SoD in E-GP models are also presented in the figure. It is clearly shown in Figure 4(a & b) that the E-SGP can lead to lower RMSE and MSLL even with the same spatial input as in the E-GP model. Furthermore, the E-SGP can create a more accurate data-driven model with less spatial input even compared to the E-GP method with full dataset. Surprisingly, the errors of the data-driven model are much lower than the errors of the coarse-grid CFD models.

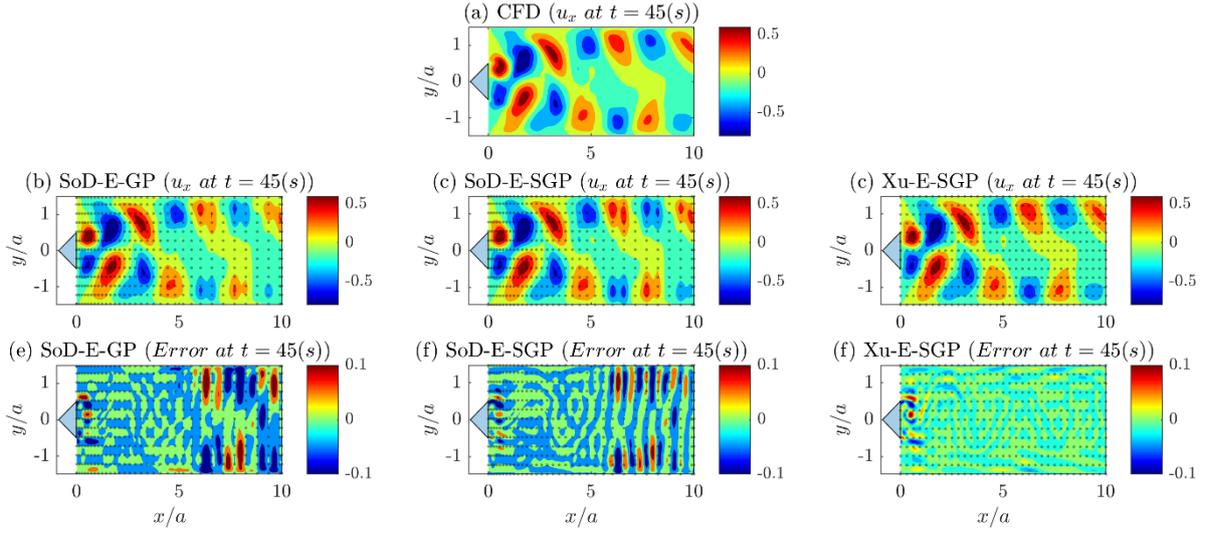

Figure 2 A single snapshot of the streamwise fluctuating velocity component ($u_x$) for the incompressible flow passing a prism bluff body produced by (a) the finest CFD simulation together with three data-driven models (b) SoD-E-GP (c) Xu-E-SGP with spatial resolution as $13 \times 33$; (d) the error contour of the prediction by SoD-E-GP model and (e) the error contour of the prediction by Xu-E-SGP model.

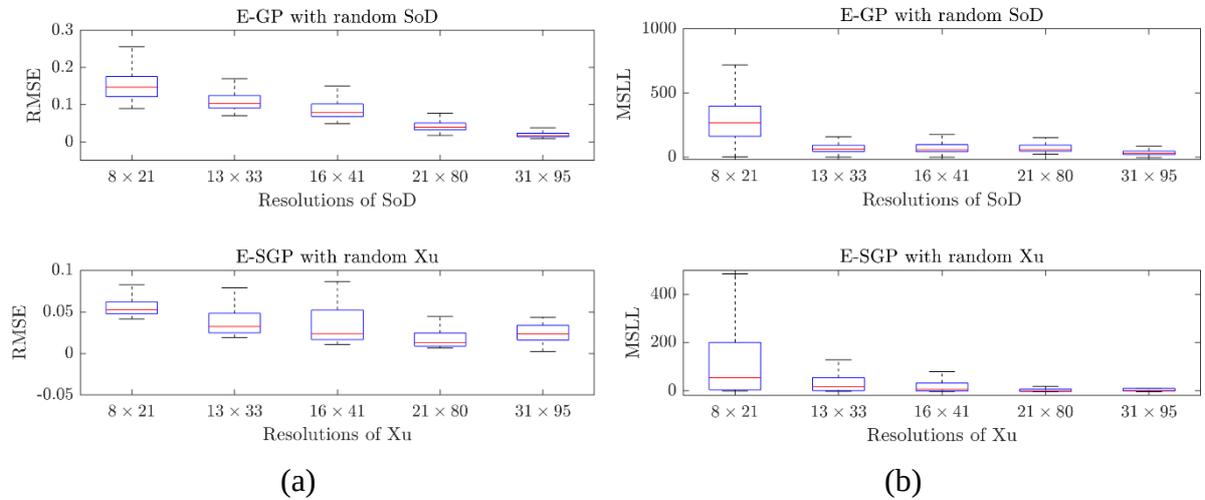

Figure 3 (a)The box plot of the rooted mean square error (RMSE) and (b) the box plot of the mean standardised log loss (MSLL) against the configurations of randomly selected subset of data (SoD) in efficient GP (E-GP) and randomly defined inducing points (Xu) in efficient SGP (E-SGP).

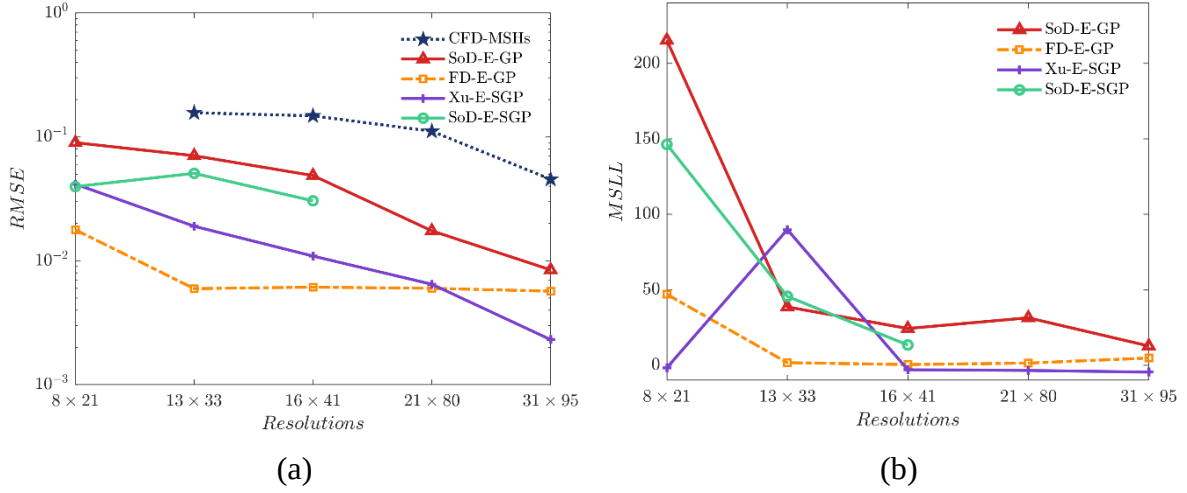

(a)    (b)

Figure 4 (a) the minimum RMSE and (b) related MSLL value against the configuration of grids for five modelling methods, namely, 1. CFD-MSHs (CFD with different meshes, only for RMSE comparison); 2. SoD-E-GP(Subset of data with Efficient GP); 3. FD-E-GP (Full training data with efficient GP and hyperparameters from SoD-E-GP methods); 4. Xu-E-SGP (random inducing data points with efficient SGP); 5. SoD-E-SGP (Efficient SGP model with Subset of data as inducing data points and hyperparameters from SoD-E-GP methods).

4.2   Lid-driven cavity flow with unstructured meshes.

The performance of the E-SGP and E-GP algorithms in reproducing unstructured fluid data are compared in this section. Here we considered the flow in a 1(m)×1(m) cavity where the flow is driven by the top moving wall, as shown in Figure 5(a). Data-driven models using E-SGP or E-GP are trained to learn the trend of the x-direction velocity component due to changes in the moving wall. The time taken to train the model (hyperparameters optimisation) and the ability of the trained model in predicting unknowns will then be assessed.

CFD models, solved using the laminar flow solver in the commercial CFD software Star-CCM+, are used to provide the training and validation databases. In the training database, $u_{wall}$ varies as 0.02 m/s, 0.04 m/s, 0.08 m/s, 0.2 m/s, 0.64 m/s, 1.0 m/s and 1.5m/s, whilst cases with $u_{wall}$ = 0.7 m/s and 0.9 m/s are treated as the validation database. The training database here is produced using the CFD model with unstructured triangular mesh containing 6437 mesh elements, as illustrated in Figure 5 (b), thus the full training dataset contains 6437 × 7 points. The validation data is generated using different meshes, a quadrilateral mesh for $u_{wall} = 0.7 m/s$ (Figure 5(c) ) and a polygonal mesh for $u = 0.9 m/s$ (Figure 5(d)). It is worth noting here that the spatial input space (i.e., locations of the data points in the flow domain) is orthogonal to the operating input space (i.e., moving wall velocity).

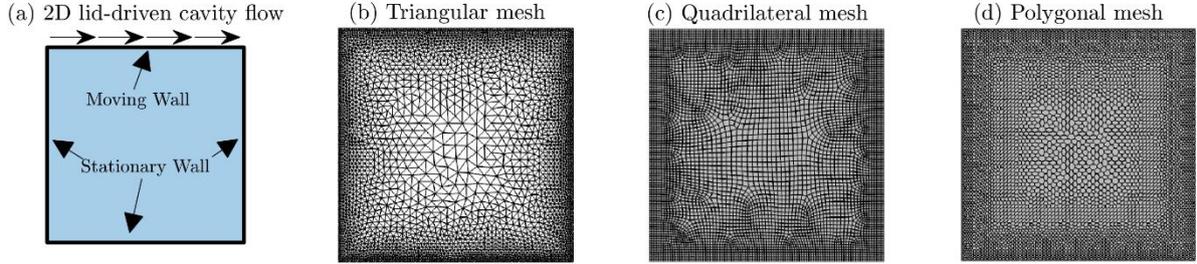

Figure 5 (a) geometry of the lid-driven cavity flow; (b) a view of the triangular mesh of the CFD model for generating the training dataset; (c) the quadrilateral mesh for the validation case of $u_{wall} = 0.7\ m/s$; (d) the polygonal mesh for the validation case of $u_{wall} = 0.9\ m/s$.

The performance of the E-GP and E-SGP algorithms on unstructured fluid datasets will be evaluated by comparing the quality of the E-GP and E-SGP models trained on 10%, 20%, 40%, 80% and 100% of the full training dataset. Furthermore, the scalability of both approaches will be examined by comparing the execution time of 1000 Metropolis-Hastings samplings (MH samplings) versus the number of training data. It is known that spatial inputs and moving wall velocities belong to two orthogonal input subspaces and the number of spatial inputs (6437) is much larger than the number of considered moving wall velocities (7). Therefore, in this example, we scale down the training data by reducing the number of spatial inputs and their related observation ($u_x$).

In the E-GP model, the full covariance matrix is the Kronecker product of two sub-covariance matrices, one for the spatial inputs and the other for the moving wall velocities. In terms of the data points in the subset of training data (SoDs), we try to distribute the data points as evenly as possible in the flow domain. An example of the SoD is shown in Figure 6 (a). The inducing points in E-SGP models are the same, and they are located on a grid, more specifically, 20 points on the x-axis, 30 points on the y-axis and 7 points (same as in the full training dataset) on the axis of the moving wall velocity. The inducing points on the xy-plane is presented in Figure 6(b). The data-driven models are trained using the Metropolis-Hastings (MH) algorithm. One hundred epochs, with 1000 MH-Samplings in each epoch, are used in the training, except for the E-GP models trained using 80% and 100% of the full training dataset. E-GP models with 80% and 100% of the full dataset scale very poorly, the training epochs for both models are set to ten.

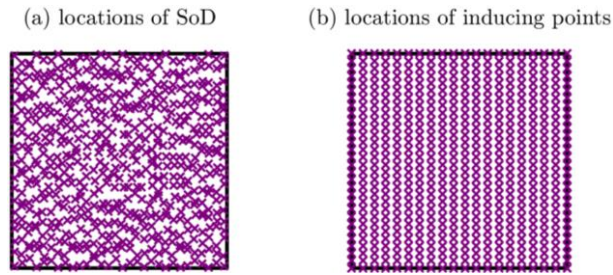

Figure 6 (a) spatial locations randomly selected subset of data for the E-GP-SoD (597 spatial points), (b) the uniformly distributed inducing points in the flow domain with resolution as 20 × 30.

Figure 7 illustrates the relationships between the execution time of 1000 MH samplings and the percentage of the full training data for the E-GP and E-SGP algorithms. It is clearly shown in the figure that the E-SGP algorithm scales much better than the E-GP algorithm. Although the E-GP model still benefits from the orthogonality of the input subspaces, the sub-covariance matrix for spatial input gradually grows with the increment of the training data. Also, the unstructured spatial input prevents the E-GP model from further accelerating the training as the E-GP for the unstructured input in Section 4.1. Unstructured spatial input is not a problem for the E-SGP, as the inducing points can be user-defined and independent from the training data. In addition, inducing points in the E-SGP models are conveniently allocated on a grid, thus making full use of the Kronecker product.

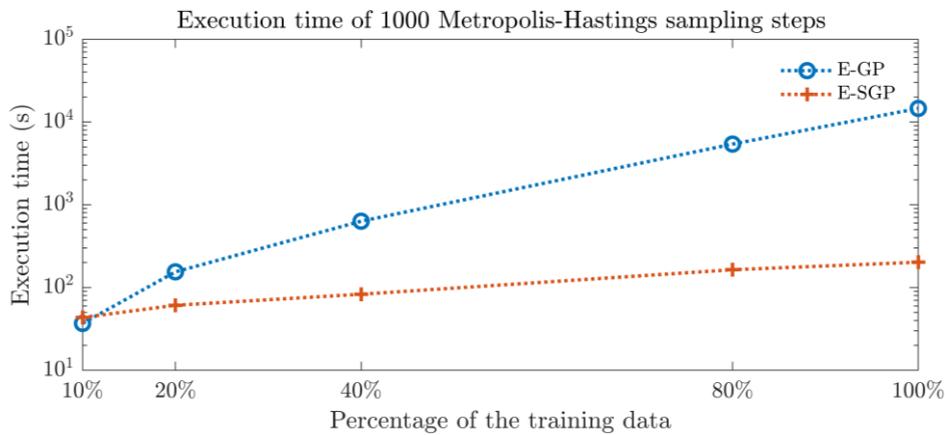

Figure 7 Execution time of the 1000 MH sampling of 'E-GP' and 'E-SGP' with 10%, 20%, 40%, 80% and 100% of full training dataset.

The statistics of the rooted-mean-square-error (RMSE) and the mean-standardised log loss (MSLL) of E-GP and E-SGP models are plotted in Figure 8 and Figure 9. Figure 8 illustrates that variances of the RMSE of E-SGP models are generally larger than those of E-GP models. This is because the resolution of the E-GP model increases with the number of training data, while the resolutions (configurations of the inducing point) in the E-SGP models are the same (30×20×7). However, the E-SGP model can produce much lower MSLL compared to E-GP models as shown in Figure 9. MSLL consists of two parts, the one is the level of model variance, the other is the ratio of the squared error and the model variance. For a high-quality Bayesian machine learning model, the MSLL should be smaller than zero and the smaller the better. However, MSLLs of all E-GP models are generally above zero, implying the model variance is either too large or too small. Considering the RMSEs of E-GP models are lower than 0.1, the high MSLL of E-GP models means that the trained E-GP models are over-confident. In other words, the model variances of E-GP models are unrealistically low. Similar observations can be found in Figure 10, which depicts the relationship between the percentage of the training data and the minimum RMSE and corresponding MSLL of the E-GP and E-SGP models. In comparison with the significant improvement in the MSLL, the RMSEs of E-SGP models are only marginally smaller than their counterparts, especially when the percentage of the full training data is larger than 40%.

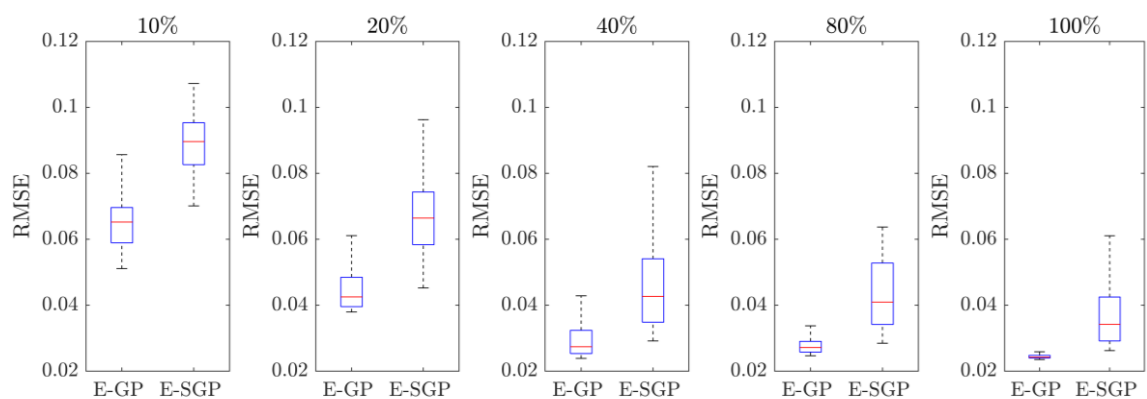

Figure 8 The box plot of the rooted mean squared error (RMSE) of E-GP and E-SGP model trained using 10%, 20%, 40%, 80% and 100% of full training dataset.

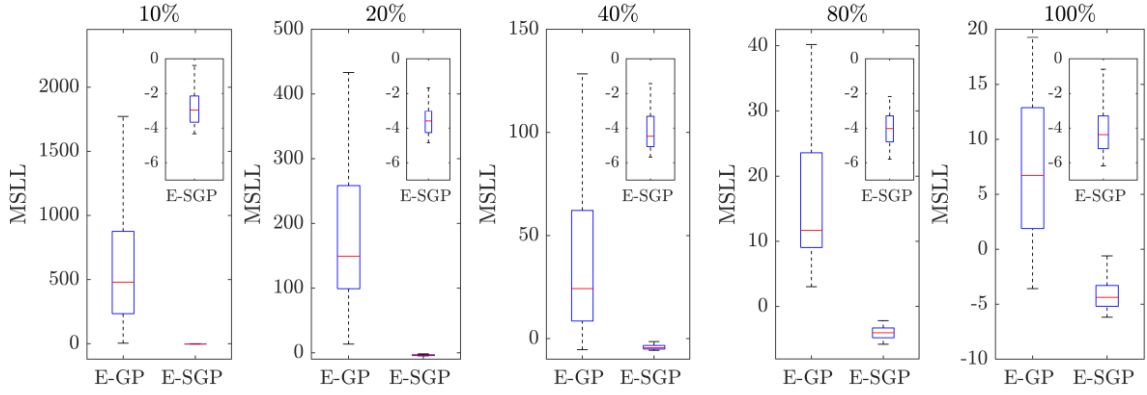

Figure 9 The box plot of the rooted mean squared error (RMSE) of E-GP and E-SGP model trained using 10%, 20%, 40%, 80% and 100% of full training dataset.

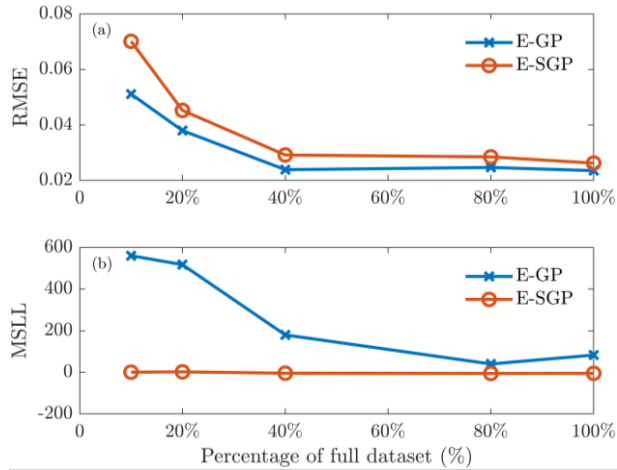

Figure 10 (a)The relationship between percentage of full training dataset with the minimum RMSE and (b) related MSLL of the model trained using E-GP and E-SGP model respectively.

## 5 Summary and discussion

In this work, we derive the Kronecker product accelerated efficient sparse Gaussian process (E-SGP) for structured and unstructured mesh fluid datasets. The novel algorithm greatly enhances the computational tractability, and efficiency, of the approximated sparse GP algorithm VEF-GP without adding extra parameters or imposing any restrictions on the prohibitively large covariance matrix. Furthermore, the separation between the input of training data and the structured inducing input allows for one to use a database with different resolutions, such as the CFD data generated using different meshes. This paper also demonstrated empirically that E-SGP outperforms E-GP in terms of scalability as well as model quality, in terms of MSLL.

In the case of similar model resolutions, i.e., the number of training data in E-GP being the same as the number of inducing points in E-SGP, the E-SGP always produces lower RMSE values. This is because E-SGP can utilise the full training dataset without significantly increasing the computational burden of solving the inverse of the covariance matrix, while E-GP only considers the subset of the full training data. When the amount of training data in E-GP and E-SGP models is the same, E-GP marginally outperforms E-SGP in terms of RMSE. Even though the resolution of the inducing data points in all E-SGP models remains the same as that of E-GP with the smallest SoD, the RMSE difference between the E-GP model and the E-SGP model gradually disappears with the increasing training data. Additionally, E-GP is also more likely to lead to unreasonably large MSLLs compared to E-SGP, in our experiments. This means that the uncertainty estimates for E-GP model predictions may be unrealistic, and such unfavourable overconfidence may cause biased conclusions in the reliability analysis of safety-critical engineering problems.

Although the major purpose of this work is to expand the application of GP with respect to different types of datasets in fluid dynamics problems, the method itself can be applied more generally to other applications. For example, it would be interesting to apply this method to generate the Bayesian ML model for interdisciplinary problems such as parametric analysis of thermal fatigue phenomena caused by thermal stripping or thermal stratification. Another important area of future research is the integration of this algorithm with other physics-informed/physics-constrained/physics-guided ML concepts to produce an efficient Bayesian ML solver for fluid dynamics problems.

## Acknowledgement

Dr. Michael Bluck and Dr. Yu Duan would like to thank the financial support from the UK Engineering and Physical Sciences Research Council (EPSRC) via the research grant entitled "ATF Solutions to Light Water-Cooled SMRs" (EPSRC grant number: EP/X011313/1).